\pdfoutput=1
\documentclass[11pt]{article}

\usepackage[]{ACL2023}

\usepackage{times}
\usepackage{latexsym}

\usepackage[T1]{fontenc}
\usepackage[utf8]{inputenc}

\usepackage{microtype}

\usepackage{inconsolata}
\usepackage{multirow}
\usepackage{hhline}
\usepackage{pifont}

\usepackage{algorithm}
\usepackage{algpseudocode}
\usepackage{amssymb}
\usepackage{amsmath}
\usepackage{mathtools}

\usepackage{booktabs}
\usepackage{xspace}
\usepackage{xcolor}
\usepackage[cjk]{kotex}
\usepackage{graphicx}
\usepackage[normalem]{ulem}
\useunder{\uline}{\ul}{}
\newcommand{\datasetname}{\texttt{HistRED}\xspace}
\newcommand{\bookname}{\textit{Yeonhaengnok}\xspace}
\newcommand{\SL}{\textit{SL}\xspace}

\newcommand{\Location}[1]{\textcolor{teal}{\textbf{#1}}}
\newcommand{\Person}[1]{\textcolor{blue}{\textbf{#1}}}
\newcommand{\Number}[1]{\textcolor{violet}{\textbf{#1}}}
\newcommand{\Product}[1]{\textcolor{brown}{\textbf{#1}}}
\newcommand{\Clothes}[1]{\textcolor{orange}{\textbf{#1}}}

\title{HistRED: A Historical Document-Level Relation Extraction Dataset}

\author{Soyoung Yang \hspace{0.3cm} Minseok Choi \hspace{0.3cm} Youngwoo Cho \hspace{0.3cm} Jaegul Choo \\
        KAIST AI\\
        \texttt{\{sy\_yang, minseok.choi, cyw314, jchoo\}@kaist.ac.kr }
        }

\begin{document}

\maketitle
\begin{abstract}

Despite the extensive applications of relation extraction (RE) tasks in various domains, little has been explored in the historical context, which contains promising data across hundreds and thousands of years.
% In response, we present \datasetname, a historical record dataset for RE tasks.
To promote the historical RE research, we present \datasetname constructed from \bookname.
\bookname is a collection of records originally written in Hanja, the classical Chinese writing, which has later been translated into Korean.
\datasetname provides bilingual annotations such that RE can be performed on Korean and Hanja texts.
% Since historical records are translated a long time after their creation, reading bilingual texts is necessary to fully understand the text; thus, our \datasetname dataset is based on a bilingual corpus. 
% Our dataset differs from other RE datasets in two aspects;
% (1) \datasetname is a bilingual RE dataset that enables RE models to utilize parallel corpus, and
In addition, \datasetname supports various self-contained subtexts with different lengths, from a sentence level to a document level, supporting diverse context settings for researchers to evaluate the robustness of their RE models.
To demonstrate the usefulness of our dataset, we propose a bilingual RE model that leverages both Korean and Hanja contexts to predict relations between entities.
Our model outperforms monolingual baselines on \datasetname, showing that employing multiple language contexts supplements the RE predictions.
The dataset is publicly available at: \url{https://huggingface.co/datasets/Soyoung/HistRED} under \href{https://creativecommons.org/licenses/by-nc-nd/4.0/}{CC BY-NC-ND 4.0} license.

\end{abstract}

\section{Introduction}

\begin{figure}[t!]
    \centering
    \includegraphics[trim={0cm 0cm 0cm 0cm}, clip=true, width=0.9\columnwidth]{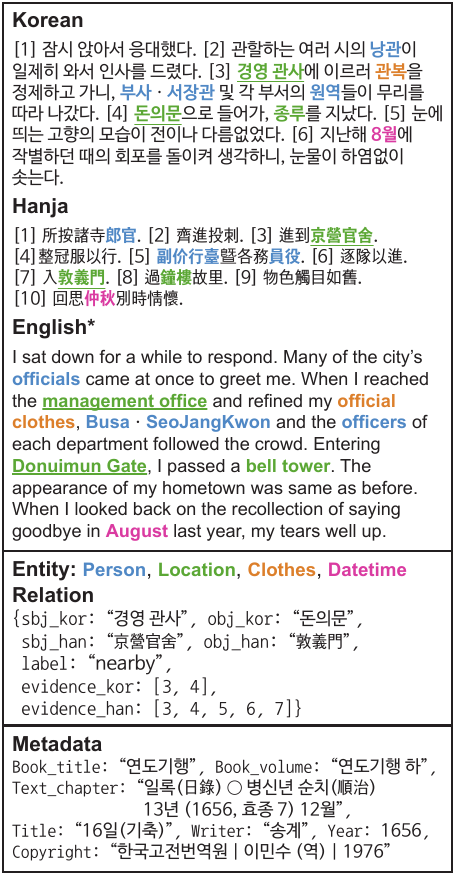}
    \caption{
    An example from \datasetname. Only one relation is shown for readability. 
    The text is translated into English for comprehension (*).
    Relation information includes (i) subject and object entities for Korean and Hanja (\textit{sbj\_kor, sbj\_han, obj\_kor, obj\_han}, (ii) a relation type (\textit{label}), (iii) evidence sentence index(es) for each language (\textit{evidence\_kor, evidence\_han}). 
    \textit{Metadata} contains additional information, such as which book the text is extracted from. 
    }
    \label{fig:example}
\end{figure}

% \usepackage{multirow}

% \begin{table*}[t]
% \center
% \begin{tabular}{cc|cc|cc|ccc}
% \multirow{2}{*}{\textbf{Data name}} & \multirow{2}{*}{\textbf{Language}} & \multicolumn{2}{c|}{\textbf{Dataset type}} & \multicolumn{2}{c|}{\textbf{Input text type}} & \multirow{2}{*}{\textbf{\# Doc.}} & \multirow{2}{*}{\textbf{\# Sent.}} & \multirow{2}{*}{\textbf{\# token in a doc}} \\ \cline{3-6}
%  &  & \multicolumn{1}{c|}{\textbf{Historical}} & \textbf{Relation extraction} & \multicolumn{1}{c|}{\textbf{Sentence}} & \textbf{Document} &  &  &  \\ \hline
% I.PHI & Ancient Greeks &  &  &  &  &  &  &  \\
% \begin{tabular}[c]{@{}c@{}}DocRED\\ (human)\end{tabular} & English &  &  &  &  & 5,051 & 40,276 & 229.64 \\
% \begin{tabular}[c]{@{}c@{}}DocRED\\ (distant)\end{tabular} & English &  &  &  &  & 101,873 & 828,115 & 231.34 \\
% KLUE-RE & Korean &  &  &  &  & 40,235 &  & 60.50 \\
% Ours & Korean &  &  &  &  & 5,862 & 8,035 & 100.57 \\
%  & Hanja &  &  &  &  & 5,862 & 23,803 & 63.96

% \end{tabular}
% \end{table*}

%%%%%%%%%%%%%%%%%%%%%%%%%%%%%%%%%%%%%%%%%%%

\begin{table*}[t]
\center
\begin{tabular}{cc|cc|cc|ccc}
\toprule
\multirow{2}{*}{Dataset} & \multirow{2}{*}{Language} & \multicolumn{2}{c|}{Dataset type} & \multicolumn{2}{c|}{Input level} & \multirow{2}{*}{\# of Doc.} & \multirow{2}{*}{\# of Sent.} & \multirow{2}{*}{\begin{tabular}[c]{@{}c@{}}\# of Tok. \\ (avg.)\end{tabular}} \\
 &  & \multicolumn{1}{c}{Historical} & Relation & \multicolumn{1}{c}{Sent.} & Doc. &  &  &  \\ \hhline{==|==|==|===} 
I.PHI & \begin{tabular}[c]{@{}c@{}}Ancient \\ Greeks\end{tabular} & 
\ding{52} &  &  & 
\ding{52} & - & - & - \\ \hline
DocRED-h & \multirow{2}{*}{English} & \multirow{2}{*}{} & \multirow{2}{*}{\ding{52}} & \multirow{2}{*}{} & \multirow{2}{*}{\ding{52}} & 5,051 & 40,276 & 229.64 \\
DocRED-d &  &  &  &  &  & 101,873 & 828,115 & 231.34 \\ \hline
KLUE-RE & Korean &  & \ding{52} & \ding{52} &  & 40,235 & 40,235 & 60.50 \\ \hline
\multirow{2}{*}{\shortstack{HistRED\\(Ours)}} & Korean & \multirow{2}{*}{\ding{52}} & \multirow{2}{*}{\ding{52}} & \multirow{2}{*}{\ding{52} } & \multirow{2}{*}{\ding{52}} & \multirow{2}{*}{5,816} & 8,035 & 100.57 \\
 & Hanja &  &  &  &  &  & 23,803 & 63.96 \\
\bottomrule
\end{tabular}
\caption{Dataset comparison. I.PHI is a dataset used for training Ithaca~\cite{asssome2022ithaca}.
DocRED-h~\cite{yao2019DocRED} is human-annotated, while DocRED-d is generated by distant supervision. 
% Our dataset statistics are shown when the sequence level (\textit{SL}) is 2.
\textbf{Historical} indicates that the dataset contains historical contents, and \textbf{Relation} means the dataset is built for the RE task. \textbf{Input level} is whether the input sequence is a single sentence (Sent.) or multiple sentences (Doc.). 
\textbf{\# of Doc.} represents the number of documents, \textbf{\# of Sent.} is the number of sentences, and \textbf{\# of Tok.} is the average number of tokens in a document using the mBERT tokenizer.}
\label{tab:data_comparison}
\end{table*}

% Itaca등이 있지만, 이 모델은 raw text base로 prediction task 수행 (누락된 글자 복원, 장소 및 시기 예측). 
% 수백 수천 년간 쌓인 수많은 역사 기록은 우리 앞에 놓인 (개인과 사회) 문제를 푸는 실마리를 제공한다.
% 하지만, 
% 우리 앞에 닥친 문제를, 수백 수천년 간 쌓인 많은 역사 기록을 바탕으로 문제 해결에 도움을 받을 수 있다. 하지만, 그 정보가 방대하여 사람이 모두 처리하기 어렵다 (labor-intensive). 따라서 여기에서 information extraction approach를 적용하여 정보 추출을 빠르게 할 수 있다. 
% Comprehending the underlying relationships in situations is often necessary for making precise and prompt decisions.
% % To make precise and prompt decisions, comprehending the underlying relationships in our situations is often necessary.
% % It is necessary to comprehend the underlying relationships in the situations we face in order to make precise and prompt decisions.
% However, as the volume of data increases rapidly, accurately analyzing data in a limited amount of time using only humans' cognitive abilities becomes increasingly challenging. % 날려?
% % it becomes challenging to analyze data accurately in a limited amount of time using only human cognitive abilities.
% % As a result, 
% Consequently, a large number of studies have been conducted to share intellectual labor with machines~\cite{Zhang_Yang_Zhao_2021}.
Relation extraction (RE) is the task of extracting relational facts from natural language texts.
To solve RE problems, diverse datasets and machine learning (ML) methods have been developed. % to solve RE problems.
Earlier work limits the scope of the problem to sentence-level RE, in which the task is to predict a relationship between two entities in a single sentence~\cite{doddington-etal-2004-automatic,walker-ace2005,hendrickx-etal-2010-semeval,alt2020tacred, stoica2021retacred}.
% various methods~\cite{wu-etal-2017-sentre, han-etal-2018-sentre} have achieved state-of-the-art performance on a sentence-level RE tasks.
% However, since sentence-level RE is hard to be applied when a large number of relational facts exist across multiple sentences.
However, such a setting is impractical in real-world applications where relations between entities can exist across sentences in large unstructured texts.
% However, because the sentence-level RE suffers in practice when a large number of relational facts exist in multiple sentences,
Therefore, document-level RE datasets for general and biomedical domains have been introduced~\cite{li-2016-cdr,yao2019DocRED,Wu2019RENETAD,klim-2021-DWIE,Luo_2022-BioRED}, serving as benchmarks for document-level RE models~\cite{docred-huguet-cabot-navigli-2021-rebel-relation, docred-tan-etal-2022-document,docred-xiao2021sais, docred-xie2021eider, docred-xu2021entity}.

Despite the vast amount of accumulated historical data and the ML methods available for extracting information from it, research on information extraction targeting historical data has been rarely conducted.
% 하지만, 방대한 역사 데이터가 쌓이고, 이를 활용할 수 있는 IE 기술이 많이 발달했음에도 불구, 역사 코퍼스를 대상으로 하는 정보 추출 연구는 거의 이뤄지지 않았다. 
% A historical domain contains promising data that can alleviate the decision-making problem along with the assistance of ML, especially in the RE problem setting.
% History helps us make better decisions in the present than in the past by learning a lesson or wisdom from historical figures or events.
% Here, the fact that the result that occurred in the distant past can affect the present makes history fascinating.
% However, it is extremely complicated to track and analyze the vast historical records covering hundreds and thousands of years.
We believe this is due to the high complexity of analyzing historical records which are written in early languages and cover hundreds and thousands of years.
% It is highly complicated to track and analyze the vast amount of historical records covering hundreds and thousands of years. 
% 현대에 사용되는 언어가 아니기때문에, 데이터를 annotate하는 것이 domain expert가 아니면 어렵다. 또한 고전 텍스트를 볼 때에는 번역본과 원문을 함께 보아야 한다.
For instance, early languages pose a challenge for accurate translation and knowledge extraction due to their differences in expressions, styles, and formats compared to contemporary languages.
Also, since historical records are translated a long time after their creation, reading bilingual texts is necessary to fully understand the text.
Such discrepancy requires domain experts who are able to understand both languages in order to accurately annotate the data.
% Therefore, historians suffer from the above problems, and there has been a consensus to utilize the AI algorithms in their corpus.
There has been a demand from historical academics to utilize ML algorithms to extract information from the huge amount of records; however, because of the aforementioned challenges, the historical domain has been overlooked by most ML communities.

% \sy{왜 document-RE 태스크가 역사 다큐먼트에 적합한가-통계 추가}
In response, we introduce \datasetname, a document-level RE dataset annotated on historical documents for promoting future historical RE studies.
\datasetname contains 5,816 documents extracted from 39 books in the \bookname corpus (see Section~\ref{sec:2} for details).
% 데이터셋 정보
% As shown in Fig.~\ref{fig:example},
% since a Hanja sentence is usually shorter than that of Korean, the informative tokens are scattered within the text.
% As a result, at least 40.5\% of the relational facts in \datasetname on the Hanja text can only be extracted from multiple sentences, while 18.56\% on the Korean.
% If we also exclude the ``per:position\_held'' relation type, 98\% of which appear in a single sentence, 55.6\% of relational facts on the Hanja exist inter-sentence, and 26.4\% on the Korean.
% This reveals that the document-level RE is an appropriate approach to our historical corpus.
% This explains why the document-level RE is appropriate approach to our \bookname corpus. 
% If we consider 
% Similar to the previous document-level dataset~\cite{yao2019DocRED}, our dataset often requires understanding the context of the corresponding document in order to predict relationships between entities in the document.
As described in Table~\ref{tab:data_comparison}\footnote{The statistics of our dataset is calculated when \SL is 2.}, our dataset is the first dataset that extracts relational information from the historical domain and
differs from other RE datasets in that it supports both sentence-level and document-level contexts, as well as two languages: Korean and Hanja.
Furthermore, researchers can select different sequence levels (\SL), which we define as a unit of context lengths, when evaluating their RE models.
% the addition of a sentence(s) back and forth from the current context.
Such independent subtexts are constructed by considering evidence sentences, which the annotators have tagged. % to each relation.
The intuition is that evidence sentences, which provide context for deriving a certain relation between two entities, should not be separated from the original text when splitting a document; thus, we introduce an algorithm that properly splits a full document into several self-contained subtexts.
% This is possible because the annotators have first tagged not only the entities and relations, but also the evidence sentence for each relation on the raw text. % by indexing the evidential sentence(s).
Finally, we propose a novel architecture that can fully utilize bilingual contexts using pretrained language models (PLMs). Experimental results demonstrate that our bilingual RE model outperforms other monolingual ones.

Our contributions are summarized as follows:

\begin{itemize}
    \item We introduce \datasetname, a historical RE dataset built from scratch on \bookname, a historical record written between the 16th and 19th centuries. 
    \item We define new entity and relation types fit for our historical data and proceed with the dataset construction in collaboration with domain experts.
    % to be appropriate with the corpus with close cooperation with experts. 
    \item We introduce a sequence level (\SL) as a unit of varying sequence lengths, which properly splits a full document into several independent contexts, serving as a testbed for evaluating RE models on different context lengths.
    
\end{itemize}

\section{Dataset Construction} \label{sec:2}
To the best of our knowledge, \datasetname is the first RE dataset in the historical domain; 
thus, there is no consensus regarding the dataset construction process on the historical corpus.
% Accordingly, there is the lack of consensus regarding data schema, document collection or annotation, and data preprocessing for historical corpus.
% dataset construction process for historical corpus.
In the process of designing our dataset, we collaborated with experts in the linguistics and literature of Hanja to arrive at a consensus.
This section describes how we collaborated with the domain experts to construct \datasetname without losing annotation quality.

\subsection{Background}
% \yw{여기 한 번 위아래랑 호응하는지 체크하기}
% In this work, we address the history of the Joseon dynasty, the former country of Korea.
Joseon, the last dynastic kingdom of Korea, lasted just over five centuries, from 1392 to 1897, and many aspects of Korean traditions and customs trace their roots back to this era.
Numerous historical documents exist from the Joseon dynasty,
% Joseon has a large number of historical records, 
including \textit{Annals of Joseon Dynasty} (AJD) and \textit{Diaries of the Royal Secretariats} (DRS).
Note that the majority of Joseon's records were written in Hanja, the archaic Chinese writing that differs from modern Chinese, because the Korean language had not been standardized until much later.
% The majority of Joseon's records are written in Hanja, which is a unique language in Korea classics and differs with modern Chinese in grammar and vocab. %\sy{한자 특징 서술}.
% the linguistics and literature of which focuses primarily on the language and history of the ancient nation of Korea. % and Hanja linguistics and literature focuses primarily on the language and history of ancient nation of Korea.
% Most of Joseon's records are written in \textit{Hanja}, and a \textit{linguistics and literature of Hanja} mainly studies the language and the history of Joseon.
We considered a number of available historical texts and selected \bookname, taking into account the amount of text and the annotation difficulty.
% 텍스트의 양, 어노테이트의 난이도를 고려하여, 
% 연행록 안에 relation 정보가 rich하기 때문이다. 
% Before constructing the dataset, we examined various texts and chose \bookname.
\bookname is essentially a travel diary from the Joseon period. %연행록은 간단히 말해서 여행일기다. 
In the past, traveling to other places, particularly to foreign countries, was rare.
% In the past, traveling to other locations, especially to foreign countries, was a rare event.
% Accordingly, the intellectuals who travel to Chung recorded their journeys in detail, and \bookname is a collection of such records. 
Therefore, intellectuals who traveled to Chung (also referred to as the Qing dynasty) meticulously documented their journeys, and \bookname is a compilation of these accounts.
Diverse individuals from different generations recorded their business trips following similar routes from Joseon to Chung, focusing on people, products, and events they encountered.
The Institute for the Translation of Korean Classics (ITKC) has open-sourced the original and their translated texts for many historical documents, promoting active historical research\footnote{The entire documents were collected from an open-source database at \url{https://db.itkc.or.kr/}}.
% Across the different generations, diverse people wrote their business trips on a similar route from Joseon to Chung and mainly described the people, products, and events they met.
% It is suitable for RE dataset construction that the contents of diverse texts are similar to each other.
% This text is suitable for building an RE dataset as it is comparable to that of diverse texts. %that the contents of diverse texts are comparable.

\subsection{Dataset Schema}
\label{sec:dataset_schema}
We engaged in rounds of deliberate discussions with three experts who have studied the linguistics and literature of Hanja for more than two decades and defined our dataset schema.
% we defined 10 entity types and 20 relation types, as well as selected the most pertinent documents.
% After the wide discussion with three experts, who have studied the linguistics and literature of Hanja more than 20 years, 
% we defined 10 entity and 20 relation types and selected appropriate documents.
% The named entities are tagged via character-level BIO (begin-Inside-Outside) tagging scheme.

\paragraph{Documents}
% As aforementioned, we choose \bookname. 
Written between the 16th and 19th centuries, the books in \bookname have different formats and contexts depending on the author or the purpose of the book. % and describe what envoys saw and how they communicated with officials of Chung, the ancient country of China.
% Because % Due to the fact that 
% the format and context vary depending on the author or the purpose of the book, it is difficult to include every book in our dataset.
After consulting with the experts, a total of 39 books that contain rich textual information were selected for our dataset, excluding ones that only list the names of people or products.
% we select 39 books as our dataset corpus, 
The collection consists of a grand total of 2,019 complete documents, with each document encompassing the text for a single day.
This arrangement is made possible because each book separates its contents according to date, akin to a modern-day diary.
% Nevertheless, some documents have an exceedingly long body of text, and thus, we split them effectively using \SL (see Section~\ref{subsec:manipul_SL}).

\paragraph{Entity and Relation Types}
% \sy{KLUE cite: "There are 30
% relation classes that consist of 18 person-related relations, 11 organization-related relations, and no_relation. Detailed
% explanation of these classes are presented in Table 10. We evaluate a model using micro F1 score, computed after
% excluding no_relation, and area under the precision-recall curve including all 30 classes."}
Since \bookname is a unique record from the Joseon dynasty, entity and relation types used in typical RE tasks are not fit for our dataset.
After conferring with the experts, we newly define the entity and relation types appropriate for our historical data.
% 역사코퍼스에 맞게 적합한 객체, 관계 타입을 정의했다. 
% the named entities and relation types must be newly defined to extract as much information as possible.
% we must define named entity and relation types that can extract as much information as possible. 
% Therefore, we named the types after extensive consultation with experts. 
% The Appendix describes these types in detail.
% The types are described in the Appendix in detail.
% by deeply coordinating with three professors in linguistics and literature of Hanja. 
The details are described in Appendix~\ref{appendix-define-schema}.

\subsection{Annotate and Collect}
\label{sec:annotate&collect}
% \sy{message of part: 전문적인 어노테이터를 모집한 뒤 충분한 보상을 지급하여 고퀄리티의 데이터를 제작하였다.}
% 한국에서 한문학을 전공하고 있는 준 석-박사 과정생 15명을 모은 뒤 전체 5주간 data 구축 작업을 진행했다. 한문과 이를 번역한 한국어 문장을 이해할 수 있고 병렬관계를 쉽게 찾아낼 수 있는 정도의 한문 이해력을 가지고 있는 학생을 우선 선발하였다. 
% 약 30\%가 박사과정, 27\% male, 73\% female. 
% 크게 두 단계로 진행하였다. 1. 총 20명의 annotator가 3주간의 기간동안 약 # 개의 텍스트를 작업. 2. annotator가 cross-check. 각 단계 전 2시간의 교육을 진행했으며, annotator가 주는 질문에 대한 대답을 실시간으로 전원에게 제공함. 
% 한 명의 어노테이터가 1차와 2차에서 작업하는 텍스트가 다르므로, 두 텍스트의 양을 합쳤을 때 모든 어노테이터의 작업량이 균둥하도록 세팅하였다. 한 어노테이터 당 평균 1700\$를 지급받았다. 
\paragraph{Annotators}
15 annotators were recruited, who can comprehend the Hanja texts with the Korean translations and have studied the linguistics and literature of Hanja for at least four years. 
% We recruit 15 annotators who can understand the Hanja text and its translated Korean text and also have studied the linguistics and literature of Hanja more than four years. 30\% of annotators are in a doctorate and 65\% are in a master's degree. 
% We payed each annotator about 1,700\$ in average. 

\paragraph{Data Annotation}
The annotation process was divided into two steps: Each annotator first annotates the text from scratch, and then a different annotator cross-checks the annotations. % they cross-check the annotated result. 
Prior to each step, we provided the annotators with guidelines and promptly addressed any inquiries they had throughout the annotation process.
The annotators were instructed to tag four types of information: entities, relation types, coreferences, and evidence sentences.
Entities are annotated in both Korean and Hanja texts, whereas the relations between entities are tagged in the Korean text only, reducing redundant workload for the annotators.
Coreferences, which are words or expressions that refer to the same entity, are also tagged such that they are all used to represent a single entity during model training.
Evidence sentences, which provide context why the entities have a particular relation, are labeled as well, following \citet{yao2019DocRED}.
For 2,019 parallel texts, the average number of sentences is 24, and the average number of characters in a sentence is 45 in Korean, and 65 and 7 in Hanja, respectively.
% The named entity and the relationship between entities in Korean text were annotated in 
% Initially, named entities in Korean and Hanja were annotated in parallel texts, and the relations between entities were tagged in Korean text only. Based on the entity annotations, the annotators tagged the relations between entities, as well as the coreferences of entities.
% We annotate the named entity in Korean and Hanja, as well as the relationship between entities in Korean text, and we conclude by adding the correlative type.
% Additionally, the annotators are instructed to identify evidence sentence(s) for each relation in both Korean and Hanja texts, following ~\citet{yao2019DocRED}.%, as our dataset is an RE dataset at the document level.

\paragraph{Preprocessing}
% \sy{message of part: 어노테이터는 전체 맥락을 보고 객체와 관계를 태깅하였고, 이를 모델 학습에 용이하게 편집하였다.}
% 우선 텍스트에서 1) exclude로 태깅된 한시와 고서 인용문 - 불필요한 부분을 제외한 뒤
% 2) 한국어 문장같은 경우에는 한문 병기 표현 제외 예; "양소영 (양소영_한자)" -> "양소영"
% 결과적으로 2,019개의 텍스트에서 5,862개의 관계 추출 데이터 인스턴스를 추출하였고, 한 인스턴스, 즉 하나의 다큐먼트의 길이는 한국어 기준 평균 24개 문장에서 2개 장, 한문 기준 평균 65개 문장에서 5개 문장으로 줄었다. 
% 위와 같이 데이터셋을 처리한 이유는 크게 두가지이다. 
% 첫 째, 한 다큐먼트의 텍스트가 길고 관계가 없는 문장이 다수 포함된 경우 인풋 시퀀스 길이에 비해 relation GT의 개수가 적어 loss가 작아지고 모델 학습에 어려움을 겪는다.
% 둘 째, 연행록을 작성할 때 하루치의 텍스트의 최소/최대 길이가 정해져있지 않아 raw text의 경우 다큐먼트 당 문장 수의 분산이 한국어 기준 1,503 (한문 12,812)었다. 하지만 위와 같은 전처리를 거친 후 한국어 4.15, 한문 57.47로 분산을 효과적으로 줄임으로써 다큐먼트의 길이를 균등히 하였다. 
The initial annotated data is preprocessed to facilitate model training due to several issues it presents.
First, some texts contain quotes from other books and poems, which may be unnecessary information for performing the RE task, and thus we exclude them from our dataset.
Second, the annotators have found no relation information in some texts either because they were too short or the author of the text had not written any meaningful information. We filter out such texts accordingly.
Lastly, the average number of sentences is quite high, with a high variance of 1,503 characters in Korean and 12,812 characters in Hanja.
This is because the writing rule of \bookname is not stringent.
% We also notice that important entity names, such as the name of an emperor, are written in Hanja next to the Korean text. Such information is helpful when reading the text 
% First, the average number of sentences in the raw text is quite high, which is 24 in Korean and 65 in Hanja. 
% Second, the raw text shows a high variance in text length, with 1,503 characters in Korean and 12,812 characters in Hanja. This is because the writing rule of \bookname is not stringent.
% as a rule for text size, when writing \bookname is not stringent.
% Third, there are sentences that lack a relation between the named entities.
% Therefore, sentences that are labeled with relations or as evidence are selected, and the entity and relationship that miss the parallel relationship between Korean and Hanja are removed. 
% After preprocessing, a total of 5,862 data instances are generated from the original 2,019 texts.
% We filtered the texts that lack relation and any parallel relation between Korean and Hanja entities.
Therefore, we divide these texts with respect to different sequence levels, as described in Section~\ref{subsec:manipul_SL}.
Consequently, the original 2,019 texts yield a total of 5,852 data instances\footnote{When \SL is 0. The detailed statistics are in Table~\ref{tab:statistics}.}.
The mean and the variance of the number of sentences are reduced from $24{\scriptstyle(1503)}$ to $2{\scriptstyle(4.15)}$ in Korean and from $65{\scriptstyle(12812)}$ to $5{\scriptstyle(57.62)}$ in Hanja. 

% We preproces the raw dataset to facilitate the model training. There are three issues on the raw dataset. First is that the average number of sentence of a raw text is very long, e.g., 24 in Korean and 65 in Hanja. Also, the variation of documents in the raw text is 1,503 in Korean and 12,812 in Hanja, because the rule fo text size when writing \bookname is not strict. Lastly, there are sentence that do not contain relation between the named entities.
% Therefore we extract sentences that have relation or tagged as evidence sentence. At this time, we remove entity and relation that omit the parallel relation between Korean and Hanja. As a result, from 2,019 long texts we generate 5,862 data instances and decrease the mean and variance of the number of sentence in a data instance from $24\pm1503$ to $2\pm4.15$ in Korea and from $65\pm12812$ to $5\pm57.47$ in Hanja. 

\paragraph{Statistics of \datasetname}
The collected dataset is split into the training, validation, and test sets, and their statistics are demonstrated in Table~\ref{tab:statistics}. 
Since the sequence length of each document varies, we first sort all data by Korean character lengths, followed by random sampling in a 2:1:1 ratio for the training, validation, and test sets, respectively.

\subsection{Sequence Level}
\label{subsec:manipul_SL}
% We introduce  as a parameter of our dataset. 
A length of a document is a major obstacle to training a PLM such as BERT, which can take sequences of length only up to a specified length, e.g., 512 tokens.
Naively, we can split long documents into multiple chunks; however, a problem may arise when the context for identifying a certain relation exists in a different chunk of text.
To resolve this issue, we introduce a sequence level (\SL), a unit of sequence length for extracting self-contained subtexts without losing context information for each relation in the text. 
This is achieved since we have instructed the annotators beforehand to mark evidence sentence(s), which are contextual sentences that help identify the corresponding relation.
As a result, we can utilize these sentences as indicators when varying the lengths of a document.

Formally, let $T^k_a$ represent a subtext for relation A when \SL is $k$.
Assume two relations exist in separate sentences of a document, i.e., $D = [s_1, \cdots, s_n]$, which consists of $n$ sentences.
When \SL is 0 and $i+1 < j$, the two subtexts can be defined as $T^0_a = [s_i, s_{i+1}], T^0_b = [s_j]$, where relation A exists in $s_i$ and its context in $s_{i+1}$, while relation B exists and has its context in $s_j$.
If \textit{SL} is set as $k$, each subtext is expanded to $T^k_a = [s_{i-k}, \cdots, s_{i+k}], T^k_b = [s_{j-k}, \cdots, s_{j+k}]$, where $1 \leq i-k$, $1 \leq j-k$, $i+k \leq n$, and $j+k\leq n$.
% , where $1 \leq i-k, j-k$ and $i+k, j+k \leq n$. 
Note that the expansion is based on the sentence where the relation exists, i.e., $s_i$ and $s_j$.
If $i-k < 1$ or $j-k<1$, we set the initial index of $T^k$ as $1$, and if $n < i+k$ or $n < j+k$, we set the last index of $T^k$ as $n$.

In addition, we must verify whether duplication occurs between the subtexts.
If $s_{i+k}$ of $T^k_a$ becomes the same sentence as $s_{j-k}$ of $T^k_b$, we combine two subtexts to a new subtext $T^k_{a+b}$ to remove the duplication between them.
As shown in Table~\ref{tab:statistics}, the size of the dataset decreases as \SL increases due to the removal of duplication.
Based on this process, we produce five versions of our dataset, where $\{0, 1, 2, 4, 8\} \in k$. 
Because our dataset contains the bilingual corpus, the new documents are first generated in Korean text, followed by constructing the corresponding Hanja subtexts.
% before modifying the Hanja text, based on the Korean text and their parallel entity relationships.
% Note that we also add any existing entities and relations to the new document. 
% In our pre-manipulated dataset

% \usepackage{graphicx}
\begin{table}[t]
\centering
\resizebox{0.75\columnwidth}{!}{%
\begin{tabular}{c|c|ccc}
\toprule
\textit{SL} & Total & |Train| & |Valid| & |Test| \\ \midrule
0 & 5,852 & 2,926 & 1,463 & 1,463 \\
1 & 5,850 & 2,925 & 1,463 & 1,462 \\
2 & 5,816 & 2,908 & 1,454 & 1,454 \\
4 & 5,704 & 2,852 & 1,426 & 1,426 \\
8 & 5,331 & 2,665 & 1,333 & 1,333 \\ \bottomrule
\end{tabular}%
}
\caption{Statistics of \datasetname}
\label{tab:statistics}
\end{table}

% \begin{tabular}{lllll}
% \toprule
% \textit{SL} & \textbf{|Train|} & \textbf{|Valid|} & \textbf{|Test|} & \textbf{Total} \\ \midrule
% 0 & 2,931 & 1,466 & 1,465 & 5,862 \\
% 1,2,4,8 & 2,925 & 1,463 & 1,463 & 5,851 \\ \bottomrule
% \end{tabular}%
% }
% \caption{Statistics of \datasetname}
% \label{tab:statistics}
% \end{table}

\section{Data Analysis}
% \sy{소영, 민석 | DocRED 참조해서 우선 적어둠}
% \sy{여기에 dataset comparison table + size + inter-sentence}
% \sy{통계적 정보, 실제로 어떻게 구축되었는지 시각화}

In this section, we analyze various aspects of \datasetname to provide a deeper understanding and highlight several characteristics of our historical data.
Table~\ref{tab:data_comparison} shows the properties and statistical aspects of \datasetname with three most related datasets: I.PHI~\cite{asssome2022ithaca}, DocRED~\cite{yao2019DocRED}, and KLUE-RE~\cite{park2021klue}. 
The tokenizer of mBERT~\cite{devlin-etal-2019-bert} is utilized to obtain the number of tokens in diverse languages.
\datasetname is the first dataset comprised of historical texts targeting the document-level RE task. 
There have been several studies on the historical corpus~\cite{assael2019pythica, asssome2022ithaca}; however, most RE datasets are based on a general or biomedical domain~\cite{yao2019DocRED,Luo_2022-BioRED}, making it hard to derive historical knowledge.% on those datasets.
% While the size of our dataset is not larger than other RE datasets, we build our dataset such that it allows levels of input sequence lengths, from a sentence to multiple sentences.
% Our dataset is more challenging than others, especially compared to the KLUE-RE dataset~\cite{park2021klue}, which is a Korean sentence-level RE dataset while ours is document-level. % and includes the relation of `no\_relation' between given named entities.
% 14,165 of 40,235 sentences contain `no\_relation,' indicating that 26,070 sentences contain meaningful relations.
% Our dataset includes 9,946 relational facts in 7,875 sentences when \SL is 0, except the `no\_relation', indicating that each sentence has an average of one relation. 

\paragraph{Named Entity Types}
\datasetname contains 10 entity types, including Location (35.91\%), Person (34.55\%), Number (13.61\%), DateTime (4.82\%), and Product (4.40\%)\footnote{The percentage is calculated when \SL is 1.}. 
On average, approximately 11 entities appear in a single document, with the median being 10. 
% The 5 most frequent entity types are location, person,  number, datetime, and product. 
The aforementioned types are the five most frequent entity types.
This can be explained that \bookname is a business-travel journal from Joseon to Chung; thus, the authors described whom they had met and when and where they had traveled.
The full description is in Appendix Table~\ref{tab:entity_full}.

\paragraph{Relation Types}
% As show in Table~\ref{tab:relation_full}, 
Our dataset encloses 20 relation types, including ``per:position\_held'' (32.05\%), ``nearby'' (27.28\%), ``alternate\_name'' (7.59\%), ``per:country\_of\_citizenship'' (5.35\%), and ``product:provided\_by'' (3.82\%)\footnote{The percentage is calculated when \SL is 1, same as entity.}. 
The frequent occurrence of ``per:position\_held'' can be explained by the distinctive writing style during the Joseon dynasty.
For instance, people wrote the name of another person along with their title (e.g., ``Scientist Alan Turing'' rather than ``Alan Turing.'')
People referred to each other by their titles or alternative names, such as pseudonyms because using a person's given name implied a lack of respect and courtesy.
The second most common relation is ``nearby,'' which indicates that the place or organization is located nearby\footnote{Since there were no mechanical mobilities and the diplomatic group moved with about 200 people, the authors could not move fast and usually walked inside a city.}. This demonstrates that the authors were interested in geographic information when traveling.
The full description is in Appendix Table~\ref{tab:relation_full}.

\paragraph{Varying Sequence Length}
As described in Section~\ref{subsec:manipul_SL}, the input text length can be altered via the sequence level (\textit{SL}). 
Table~\ref{tab:seq_len_var} shows a distribution of the number of tokens within a document when \textit{SL} changes.
When \SL is 1, our sequence length becomes longer than the sentence-level RE dataset, including KLUE-RE. %sentence level보다 길고
Additionally, when \SL $ \geq 4$, our dataset exceeds the length of other document-level RE datasets, including DocRED.% 일반 document-level보다 길다. %, even though the language of each document is distinct. 

\begin{table}[t]
\resizebox{\columnwidth}{!}{%
    \begin{tabular}{c|l|lll}
    \toprule
    \textit{SL} & Language & Mean & Var. & Median \\ \midrule
    \multirow{2}{*}{0} & Korean & 46.46 & 5,026 & 37 \\
     & Hanja & 31.56 & 2,729 & 24 \\ \hline
    \multirow{2}{*}{1} & Korean & 100.58 & 6,505 & 91 \\
     & Hanja & 64.01 & 3,786 & 56 \\ \hline
    \multirow{2}{*}{2} & Korean & 152.51 & 8,399 & 142 \\
     & Hanja & 97.78 & 5,148 & 89 \\ \hline
    \multirow{2}{*}{4} & Korean & 250.64 & 15,416 & 239 \\
     & Hanja & 163.29 & 10,224 & 153 \\ \hline
    \multirow{2}{*}{8} & Korean & 427.28 & 36,6410 & 420 \\
     & Hanja & 282.04 & 23,758 & 274 \\ \hline
    KLUE-RE & Korean & 60.50 & 918 & 54 \\ 
    DocRED-h & English & 229.64 & 5,646 & 209 \\
    \bottomrule
    \end{tabular}
    }
    \caption{
    Distribution of the number of tokens in a document for each dataset with various sequence levels (\textit{SL}). 
    We use mBERT tokenizer to get the number of tokens. 
    }
    \label{tab:seq_len_var}
\end{table}

\begin{table}[ht]
\resizebox{\columnwidth}{!}{%
    \begin{tabular}{c|c|c|c|c}
    \toprule
        $\mu {\scriptstyle(\sigma^2)}$ & $N_{init}$ & $N_{add}$ & $N_{del}$ & $N_{fin}$  \\ \midrule
        $E_{kor}$ & $51.3{\scriptstyle(96.6)}$  & $6.5{\scriptstyle(23.1)}$ & $2.2{\scriptstyle(15.2)}$ & $55.6{\scriptstyle(101.6)}$ \\ \hline
        $E_{han}$ & $50.62{\scriptstyle(95.6)}$ & $6.2{\scriptstyle(22.1)}$ & $2.0{\scriptstyle(13.8)}$ & $54.8{\scriptstyle(100.4)}$ \\ \hline
        $Rel$  & $4.9{\scriptstyle(11.4)}$   & $0.6{\scriptstyle(2.3)}$  & $0.4{\scriptstyle(1.9)}$  & $6.1{\scriptstyle(11.5)}$  \\ \bottomrule
    \end{tabular}
    }
    \caption{
        Annotation statistics during the data construction procedure. $E_{kor}$ and $E_{han}$ represent named entities in the Korean text and the Hanja text, respectively. $Rel$ is the number of relational triplets. $N_{init}$ is the number of annotations at the first step. $N_{add}$ and $N_{del}$ are the number of addition and deletions from previous annotations after cross-checking. $N_{fin}$ is the number of final annotations.
        }
        \label{tab:process-stats}
\end{table}

\paragraph{Annotation Procedure Statistics}
% 30\% of annotators are doctoral students and 65\% are master's degree students.
Since our dataset construction consists of annotation and cross-checking steps, we summarize the statistics of this procedure.
As shown in Table~\ref{tab:process-stats}, each annotator tagged an average of 51.3 Korean entities, 50.6 Hanja entities, and 4.9 relations on each raw text.
At the cross-checking step, a different annotator added an average of 6.5 Korean entities, 6.2 Hanja entities, and 0.5 relations, while deleting 2.2 Korean entities, 2.0 Hanja entities, and 0.3 relations.
As a result, the final annotations consist of 55.6 Korean entities, 54.8 Hanja entities, and 5.1 relations for each raw text on average. 
% Based on this result, we claim that the annotators tagged the text in a consistent manner.
% The high variance of annotations are because the text length varies as mentioned at Section~\ref{sec:annotate&collect}. 

\section{Bilingual Relation Extraction Model}
% \sy{소영, 민석, Figure}

Unlike translation between modern languages, such as translation from English to Korean, historical records have been translated hundreds of years after their creation. 
As a result, the gap between ancient and present makes the translation task from Hanja into Korean difficult.
Also, the translated texts can vary across translators; thus, the domain experts read both Hanja and Korean texts to fully understand the original text.
% requiring domain experts who can read both Hanja and Korean texts to fully understand the original text.
Based on this observation, we hypothesize that understanding the bilingual text would help a model extract valuable information and design our bilingual RE model.

\begin{figure}[t]
\centering
\includegraphics[trim={0cm 0cm 0cm 0cm}, clip=true, width=0.9\columnwidth]{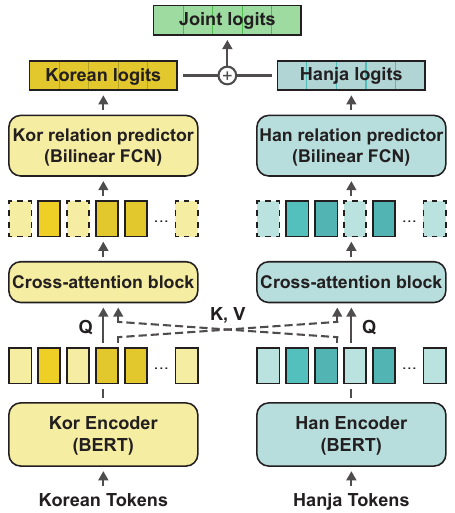}
\caption{
Architecture of our bilingual RE model.
The entities are colored in the dark compared with the other input tokens. 
``Kor Encoder'' is an encoder for the Korean language, and ``Han Encoder'' is for the Hanja language.
}
\label{fig:model}
\end{figure}

As shown in Figure~\ref{fig:model}, our model is a joint model of two separate encoders for Hanja and Korean, along with %accounting for bilingual text % is a joint model that takes bilingual text into account 
a cross-attention block from the Transformer architecture~\cite{transformer}.
For a document $D$ of length $n$ in Hanja and $m$ in Korean, we have $D_{han}=[x_t]_{t=1}^n$ and $D_{kor}=[y_t]_{t=1}^m$, where $x$ and $y$ are input tokens of each document. 
We use the PLM encoder to obtain contextualized embeddings: $H_{kor}, H_{han}$.
% \-egin{align*}
%     H_{han} &= LM_{han}([x_1, ..., x_n]), \\
%     H_{kor} &= LM_{kor}([y_1, ..., y_m])
% \end{align*}
% $$
% H_{han} = LM_{han}([x_1, ..., x_n])
% $$
% where $H_{han} \in \mathbb{R}^{n\times d}$, $H_{kor} \in \mathbb{R}^{m\times d}$ and $d$ is hidden dimension for language model.
Based on these hidden representations, we adopt the multi-head cross-attention block, which consists of a cross-attention layer and residual connection layer~\cite{transformer}.
% We get cross-attented representation $H'$ by setting key and value as another language. 
For instance, when the encoder process the Hanja text, we set the query as the Hanja token and the key and value to the Korean tokens.
Cross-attended representation $H'$ is defined as
% \begin{align*}
\begin{equation}
    H'_{han} = softmax(Q_{han}, K_{kor})V_{kor},
\end{equation}
% 
% \end{align*}
where we denote query $Q_{han} = W_Q H_{han}$, key $K_{kor} = W_K H_{kor}$, and value $V_{kor} = W_V H_{kor}$, which are all linear projections of hidden representation $H$. $W_Q\in\mathbb{R}^{d\times d}$, $W_K\in\mathbb{R}^{d\times d}$, and $W_V\in\mathbb{R}^{d\times d}$ are learnable weight matrices.
After the cross attention, $H'_{han}$ is further processed in a residual-connection layer, $Z_{han} = Linear(H_{han} + H'_{han})$.
We get $Z_{kor}$ in the same manner.
Our model pools entity embeddings from $Z_{han}$ and $Z_{kor}$.
Each bilinear classifier predicts relation types, returning separate logits: $\mathrm{logit}_{han}$ and $\mathrm{logit}_{kor} $.
At last, our model generates final logits as follows:
\begin{equation}
\mathrm{logit}_{out} = \alpha \cdot \mathrm{logit}_{han} + (1-\alpha)\cdot \mathrm{logit}_{kor},
\end{equation}
where $\mathrm{logit} \in\mathbb{R}^{k\times c} $ denotes the output logits of $k$ entity pairs for all $c$ relations, and $\alpha$ is a hyper-parameter.

% \usepackage{multirow}
% \usepackage{graphicx}
% \usepackage[normalem]{ulem}

\begin{table*}[t!]
\centering
\resizebox{\textwidth}{!}{%
\begin{tabular}{ll|ccc|ccc|ccc}
\toprule
\multicolumn{2}{l|}{} & \multicolumn{3}{c|}{\textit{SL} = 0} & \multicolumn{3}{c|}{\textit{SL} = 1} & \multicolumn{3}{c}{\textit{SL} = 2} \\
Language & Model & P & R & F1 & P & R & F1 & P & R & F1 \\  \hhline{==|======|===} 
\multirow{3}{*}{Korean} 
& mBERT & 67.80 & 58.01 & 62.53 & 66.10 & 50.63 & {\ul 57.34} & 57.43 & 42.69 & 48.97 \\
& KoBERT & 71.16 & 49.94 & 58.69 & 58.80 & 45.207 & 51.11 & 47.01 & 31.43 & 37.67 \\
& KLUE & 73.43 & 54.52 & {\ul 62.58} & 62.60 & 52.16 & 56.90 & 54.93 & 45.47 & {\ul 49.75} \\
 \hline
\multirow{2}{*}{Hanja}
& mBERT & 56.88 & 42.94 & 48.93 & 41.53 & 26.92 & 32.67 & 26.81 & 26.24 & 26.52 \\
& AnchiBERT & 63.40 & 50.04 & {\ul 55.93} & 50.28 & 32.69 & {\ul 39.62} & 32.27 & 32.12 & {\ul 32.24} \\
 \hline
% \multirow{2}{*}{Korean+Hanja}
 Korean+Hanja & Ours & 73.75 & 55.71 & \textbf{63.48} & 70.37 & 50.10 & \textbf{58.53} & 66.73 & 41.24 & \textbf{50.98}\\
 \bottomrule
\end{tabular}%
}
\caption{
Performance comparison when the sequence level (\textit{SL}) of \datasetname is 0, 1, and 2.
P, R, and F1 are precision, recall, and F1 score respectively.
All model is based on BERT-base.
All scores are described on the percentage (\%) and rounded off the third decimal point.
The \textbf{best F1 score} is in bold at each \textit{SL}, and the \underline{second} score for each language is underlined.
% Model with star (*) is\textbf{} the pretrained model on AJD and DRS from \cite{yoo2022hue}. 
}
\label{tab:performance}
\end{table*}

\section{Experiments}
% \sy{소영, 민석}

% To assess the challenges of \datasetname, we conduct experiments to evaluate the RE systems on the dataset. 

\subsection{Settings}
\paragraph{Models} 
Since our dataset consists of two languages, we build separate models for each language.
We implement all models based on Huggingface Transformers~\cite{wolf-etal-2020-transformers}.
For Korean, the baselines are mBERT~\cite{devlin-etal-2019-bert}, KoBERT (a Korean BERT)\footnote{\url{https://github.com/SKTBrain/KoBERT}}, and KLUE~\cite{park2021klue}. 
For Hanja, the baselines are mBERT and AnchiBERT~\cite{tian2020anchibert}. 
For our bilingual model, we consider combinations of these PLMs, i.e., KLUE, KoBERT, and mBERT for the Korean encoder and mBERT and AnchiBERT for the Hanja encoder.
In our experiments, the combination of KLUE and AnchiBERT shows consistent scores when varying \SL.
Therefore, our model consists of KLUE and AnchiBERT for Korean and Hanja encoders.
% The AnchiBERT* are , where AnchiBERT are pretrained on Annals of Joseon Dynasty (AJD) and Diaries of the Royal Secretariats (DRS), which are the large historical corpus written in Joseon dynasty.
% AnchiBERT is a BERT-based model which is pretrained on 39.5M tokens of ancient Chinese, i.e., Hanja and finetuned with AJD and DRS, where the finetuned model is provided by HUE~\cite{yoo2022hue} as open-source.

\paragraph{Evaluation Metric} 
% \sy{민석님. F1 score 어떻게 구했는지 작성}
Following previous work in RE~\cite{yao2019DocRED}, precision, recall, and micro-F1 scores are used for evaluating models.

% We set hyper-parameters as follows. 
\paragraph{Hyper-parameters} 
Hyper-parameters are set similarly to the BERT-base model in~\citet{devlin-etal-2019-bert}.
The size of the embedding and hidden vector dimensions are set to 768, and the dimension of the position-wise feed-forward layers to 3,072.
All encoders consist of 12 layers and 12 attention heads for each multi-head attention layer.
Also, the cross-attention block consists of 8 multi-head attention, and $\alpha$ is set as 0.5 when we get the final logits ($L_{out}$). However, when \SL is 2, 4, and 8, $\alpha$ is set to 0.6. 
The batch size for all experiments is set to 8.
The learning rate is set to 5e-5 using the Adam optimizer~\cite{adam-bengio2015}.
All models are trained for 200 epochs and computed on a single NVIDIA TESLA V100 GPU.
Computational details are in Appendix~\ref{sec:appendix_exp_compu}.

\subsection{Results}
% \paragraph{Model Performance.}
As shown in Table~\ref{tab:performance}, our model outperforms other monolingual baselines and consistently demonstrates the best performance even as \SL grows.
% Note that our model consists of KLUE and AnchiBERT. 
Even though KLUE as a monolingual model performs worse than mBERT when \SL is 1, our model, which combines KLUE and AnchiBERT, outperforms mBERT. This indicates that exploiting bilingual contexts improves performance.
%there is a distant gap between the scores of KLUE, which presents the best score between the monolingual models. 
% performance differs significantly when \textit{SL} is 0, 2, and 4. 
% Our model also maintains the best score compared with other baselines, whereas the scores of AnchiBERT drop significantly as the \SL increases.
% The stability of our model can be explained that
We believe that the cross-attention module and the joint architecture not only incorporate the knowledge from the Korean model, but also create synergy between the Korean and Hanja language models by compensating for each other's deficiencies.
We test this hypothesis with analysis in Section~\ref{sec:analysis}.
% This implies that our model not only utilizes the knowledge from the Korean model but also make synergy of Korean and Hanja language by covering each other's shortcomings. 
% Consequently, the experimental results demonstrate that a historical record can be effectively processed using the joint model if the record is translated into the contemporary language.
Consequently, the experimental results imply that utilizing a bilingual model would be efficient in analyzing other historical records if the record is written in an early language and translated into a modern one.
% The experimental results demonstrate that processing a historical record in a joint model is effective if the records have its translated text on concurrent language. 
% The detailed studies are described in Section~\ref{sec:analysis}. 

As our dataset also supports using only one language, we also make note of the monolingual performance.
In the Korean dataset, KLUE outperforms mBERT and KoBERT when \SL is 0 and 2, while mBERT performs better than KLUE when \SL is 1.
We also find that KoBERT shows worse performance than mBERT, even though KoBERT was trained specifically on the Korean corpus.
This demonstrates that our historical domain is dissimilar from the modern Korean one.
In Hanja, AnchiBERT performs best regardless of input text length.
Additional experimental results are reported in Appendix Table~\ref{tab:performance-sl-48}.

% \paragraph{Human Performance.}

\begin{figure*}[t]
    \centering
    \includegraphics[trim={0cm 0cm 0cm 0cm}, clip=true, width=\textwidth]{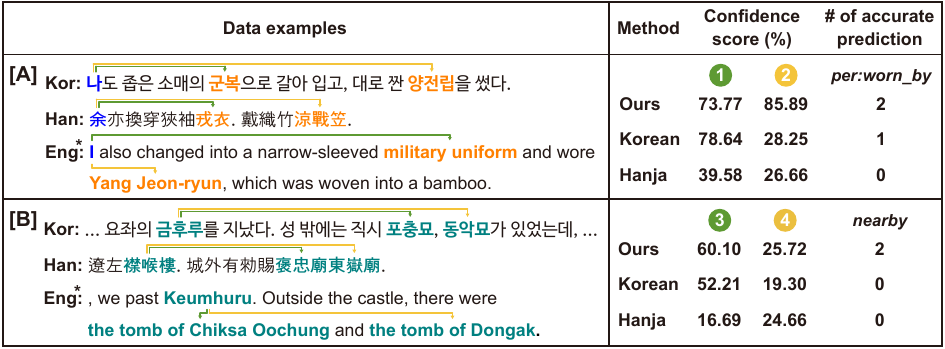}
    \caption{
    Case study of our dataset. We compare our model with two monolingual baselines: KLUE for Korean and AnchiBERT for Hanja. The bold blue represents for ``\Person{person}'' entity, orange for ``\Clothes{clothes},'' and green for ``\Location{location}.'' 
    The sentences are extracted from documents for readability, and translated into English for comprehension (*).
    % We translate the Korean text to English for these examples. 
    }
    \label{fig:case_study}
\end{figure*}

\section{Analysis}
\label{sec:analysis}

In this section, we introduce a real-world usage scenario and analyze our model on \datasetname, describing how our historical dataset can be utilized in detail.

\subsection{Usage Scenario of \datasetname}

Let us assume that a domain expert aims to collect information about the kings of Chung. 
In our dataset, he or she can extract the facts via the entity of ``Hwang Jae (황제)'' in Korean, which is a particular word to indicate the emperors of Chung, and chronologically order the events around the title. 
Note that this is possible because our dataset contains (i) the text in both Korean and Hanja and (ii) the year when the text was written. 
In total, 34 relational facts are derived from eight distinct years between 1712 and 1849, including that
(a) the king in 1713 had the seventh child via the ``person:child'' class, and
(b) the king in 1848 presented the various products with specific names, including ``五絲緞'' and ``小荷包,'' to Joseon via the ``product:given\_by'' class.
Since most of the historical records only mentioned a crown prince of Chung, describing the seventh child of the king of Chung is a rare event, which can be a motive for other creative writings.
In addition, the exact name of the products the king gives reveals that those products were produced in Chung in 1848 and would be a cue to guess the lifestyle of Chung. 
% Based on these facts, we can infer (i) the family relationship of Chung and (ii) the diplomatic ties between Joseon and Chung.

The expert can derive the facts from our dataset only by reading the 34 relational facts. 
However, if he or she has to extract them from the raw corpus, they must read at least 20 raw documents containing 1,525 sentences in Korean and 4,995 in Hanja. 
This scenario illustrates how \datasetname can accelerate the analysis process in the historical domain.

\subsection{Advantage of the Bilingual RE Model}

% Experiments are driven when \textit{SL} is 4.
% KLUE is used to process the Korean text and AnchiBERT for the Hanja, and our model consists of KLUE and AnchiBERT.
To analyze the stability of our joint model, we compare three models on random samples from the test set.
We use KLUE and AnchiBERT models independently for a monolingual setting, whereas we combine them for our joint model. The SL is set to 4.
% \subsection{Case Study on Model Prediction}
% As shown in Fig.~\ref{fig:case_study}, two data instances are sampled from test set to analyze why our model shows better performance than the monolingual models when \SL is 4.
% To analyze the reason why our model performs better than the monolingual models, we sample two data instances.
As shown in Figure~\ref{fig:case_study}, we sample two examples: case A and B, each of which displays the most representative sentences that contain the relations for the sake of readability.
% Each case contain two relation triplets, thus we will define each relation triplet as number.
% For readability, we only extract the main sentences that contains relations, and the full text is available in Appendix.
In both examples, our model successfully predicts accurate relation classes.
In the case of A, the ground truth (GT) label is ``per:worn\_by'' for first and second relation triplets. Despite the successful prediction of our model with relatively high confidence scores,
the Korean model matches only one of the two, while the Hanja model fails to predict both.
% our model predicts both relation classes with a confidence score of at least 0.6, 
% whereas the monolingual model predicts only one class, i.e., the first class on Hanja and the second class on Korean.
% It implies that our model has learned the advantages of our training method, which utilizes both corpora. %approach, which trains our model on both corpora.
In the case of B, the GT label is ``nearby'' for the third and fourth ones.
Since the third and fourth relations exist across sentences, predicting them is crucial for a document-level RE task.
Our model successfully predicts both relation types even with a low confidence score, while the other monolingual models fail. 
% although it has low confidence score at the fourth relation
% Particularly, our model accurately classifies the fourth class, while the other models predict it as a ``none'' class.
% This example shows the difficulty of understanding the text written in Hanja.
% While the original (Hanja) text has five sentences, there is only one sentence in the Korean text. 
% The Korean text is written in a single sentence, whereas the original Hanja text contains five.
% Also, the Korean shows the same phrase to indicate `a place of residence' while the Hanja represents a different character, i.e., `處焉' and `所處.' 
This case study confirms our hypothesis on our joint model; the jointly trained model can improve the performance by compensating for each monolingual model's weaknesses, and our model successfully harmonizes the separate PLMs.

\section{Related Work}
% In this section, we will briefly review the related work in two aspects.
% We compare our dataset with previous work as shown in Table~\ref{tab:data_comparison}.

\subsection{Relation Extraction}
% \sy{민석. 간단하게 한문단 정도로}
% \sy{DocRED, TacRED, KLUE}
% \sy{RE 간단하게 정의, 역사, 모델 정의}

RE datasets~\cite{yao2019DocRED, alt2020tacred, stoica2021retacred, park2021klue, Luo_2022-BioRED} have been extensively studied to predict relation types when given the named entities in text.
RE dataset begins at the sentence level, where the input sequence is a single sentence.
This includes human-annotated datasets~\cite{doddington-etal-2004-automatic,walker-ace2005,hendrickx-etal-2010-semeval} and utilization of distant supervision~\cite{reidel-2010-pkdd} or external knowledge~\cite{cai-etal-2016-bidirectional, han-etal-2018-fewrel}.
% where the input text is a single sentence.
Especially, TACRED~\cite{alt2020tacred, stoica2021retacred} is one of the most representative datasets for the sentence-level RE task.
However, inter-sentence relations in multiple sentences are difficult for models trained on a sentence-level dataset, where the model is trained to extract intra-sentence relations.
To resolve such issues, document-level RE datasets~\cite{li-2016-cdr,yao2019DocRED,Wu2019RENETAD,klim-2021-DWIE,Luo_2022-BioRED} have been proposed.
Especially, DocRED~\cite{yao2019DocRED} contains large-scale, distantly supervised data, and human-annotated data.
KLUE-RE~\cite{park2021klue} is an RE dataset constructed in the Korean language. 
However, KLUE-RE is a sentence-level RE dataset, making it challenging to apply document-level extraction to the historical Korean text. 
% so it is hard to apply document-level relation extraction tasks on Korean text.
To the best of our knowledge, our dataset is the first document-level RE dataset in both Korean and Hanja. 
% Because our document-level dataset consists of Korean and Hanja, we anticipate the document-level RE task would be constructed in Korean. 

\subsection{Study on Historical Records}
% \sy{소영, 영우}
% \sy{ITHACA, HUE, }
Several studies have been conducted on the application of deep learning models in historical corpora, particularly in Ancient Greece and Ancient Korea.
% Utilization of deep learning model on historical corpus has been studied widely, especially on Ancient Greece and Ancient Korea.
The restoration and attribution of ancient Greece~\cite{assael2019pythica,asssome2022ithaca} have been studied in close collaboration with experts of epigraphy, also known as the study of inscriptions. 
In Korea, thanks to the enormous amount of historical records from the Joseon dynasty, a variety of research projects have been conducted focusing on AJD and DRS % diverse tasks have been researched applying various models
~\cite{yang2005analysis, bak2015five, hayakawa2017long, ki2018horse, bak2018conversational, kang2019wordhanjakor, kang2021restore, yoo2022hue}.
In addition, using the Korean text of AJD, researchers have discovered historical events such as magnetic storm activities~\cite{hayakawa2017long}, conversation patterns of the kings of Joseon~\cite{bak2018conversational}, and social relations~\cite{ki2018horse}. 
\citet{kang2021restore} also suggests a translation model that restores omitted characters when both languages are used.
\citet{yoo2022hue} introduce BERT-based pretrained models for AJD and DRS. 
As interests in historical records grow, numerous research proposals have emerged.
However, most studies only utilize the translated text to analyze its knowledge.
% most studies just utilize the translated text to analyze the knowledge in the text.
In this paper, we aim to go beyond the studies that rely solely on the text. 

% \section{Discussion}
% - 

\section{Conclusion}
% \sy{소영, 민석, 영우}
% In this paper, we present \datasetname, a document-level relation extraction dataset of historical corpus based on \bookname.
In this paper, we present \datasetname, a document-level relation extraction dataset of our historical corpus.
Our study specializes in extracting the knowledge in \bookname by working closely with domain experts.
The novelty of \datasetname can be summarized by two characteristics: it contains a bilingual corpus, especially on historical records, and \textit{SL} is used to alter the length of input sequences.
We also propose a bilingual RE model that can fully exploit the bilingual text of \datasetname and demonstrate that our model is an appropriate approach for \datasetname.
We anticipate not only will our dataset contribute to the application of ML to historical corpora but also to research in relation extraction.

\section*{Limitations}
We acknowledge that our dataset is not huge compared to other sentence-level relation extraction datasets. However, \datasetname is the first bilingual RE dataset at the document level on the historical corpus. In addition, we constructed 5,816 data instances, and our bilingual model trained on \datasetname achieved an F1 score of 63.48 percent when SL is 2. This reveals that our dataset is sufficient for finetuning the pretrained language models.
% One limitation of our study is the size of our dataset. 
% Even though the number of data instances has increased from 2,019 to 5,862, this dataset is still smaller than other RE datasets.
% Although we increase the number of data instances from 2,019 to 5,862, it is still smaller than other RE datasets. 
% Also, because \bookname is a collection of travel records, the domain is not wide as other records on Joseon dynasty.
Also, because \bookname is a collection of travel records, the domain is not as expansive as other Joseon dynasty records.
Additional research on massive corpora covering a broader domain is required in future studies.

% is required on a larger corpus covering a broader domain.
% Further studies are needed on a bigger corpus by covering a wider domain.
% In future, we will expand the dataset to larger corpus, such as AJD and DRS. 
% At last, because of to the long-time history of \bookname, some characters of Hanja text is removed and typed as \square. 
% \sy{소영, 민석, 영우 | 맨 마지막에 추가 | 적는 것 필수. 없으면 데스크리젝 }
% - 본문에 오타있는 경우가 있음
% - 소실된 한자 포함
% - 소제목의 경우 뒤에 

\section*{Ethical Consideration}
% https://aclrollingreview.org/responsibleNLPresearch/

% 1.  if you collected data via crowdsourcing, did your instructions to crowdworkers explain how the data would be used?
% 2. Did you report the basic demographic and geographic characteristics of the annotator population that is the source of the data?
We conducted two separate meetings before the first and second steps of data construction.
At first, we introduced the reason we built this dataset and the goal of our study and clarified what the relation extraction task is and how the dataset will be used.
All annotators agreed that their annotated dataset would be used to build an RE dataset and train neural networks. 
We explained each type of the named entity and the relation with multiple examples and shared user guidance.
In the second meeting, we guided the annotators in evaluating and modifying the interim findings in an appropriate manner. 

We adjusted the workload of each annotator to be similar by assigning different text lengths during the first and second steps.
We compensated each annotator an average of \$1,700, which is greater than the minimum wage in Korea.
Among 15 annotators, 14 were Korean, one was Chinese, 11 were female, and four were male. 30\% of annotators are in a doctorate and 65\% are in a master's degree. 
Regarding copyrights, since our corpus is a historical record, all copyrights belong to ITKC. ITKC officially admit the usage of their corpus under \href{https://creativecommons.org/licenses/by-nc-nd/4.0/}{CC BY-NC-ND 4.0} license. 

\section*{Acknowledgement}
This research was supported by the KAIST AI Institute (``Kim Jae-Chul AI Development Fund'' AI Dataset Challenge Project) (Project No. N11210253), the National Supercomputing Center with supercomputing resources including technical support (KSC-2022-CRE-0312), and the Challengeable Future Defense Technology Research and Development Program through the Agency For Defense Development (ADD) funded by the Defense Acquisition Program Administration (DAPA) in 2022 (No. N04220080).
We also thank Junchul Lim, Wonseok Yang, Hobin Song of Korea University, and the Institute for the Translation of Korean Classics (ITKC) for their discussions and support. 
% \sy{임준철교수님, 양원석교수님, 송호빈교수님, 김연주 박사, 고전번역원에 감사하다. + 사사표기}

\bibliography{anthology,custom}
\bibliographystyle{acl_natbib}

\clearpage
% \newpage
\appendix

\begin{figure*}[t]
    \centering
    \includegraphics[trim={0cm 0cm 0cm 0cm}, clip=true, width=1.0\textwidth]{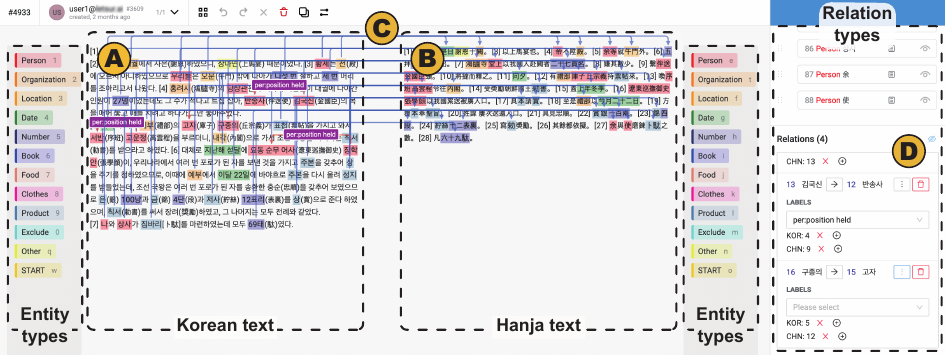}
    \caption{
    User interface for data annotation.
    We divide the overall step into four notations: A, B, C, and D.
    A, B, and C are for the entity annotation step, and D is for the relation annotation.
    The annotators detect the named entity of the Korean text in A, find the parallel entity of the Hanja text in B, and annotate the parallel relationship, shown as the blue line in C.
    After checking the entity detection, the annotators move to D, where they annotate the relation between the entities, choose the relation class, and add the indices of evidence sentences.
    }
    \label{fig:UI}
\end{figure*}

\section{Dataset Construction}

The procedure consists of the following five steps: 1) collecting corpus from the open-source data of ITKC; 2) defining the schema of the named entities and relations; 3) identifying the entities in given documents; 4) annotating corresponding relations; and 5) modifying the interim results.
This section illustrates the overall procedure.
% 다큐먼트보다 긴 텍스트에 대해서 작업을 진행하였기 때문에, 휴먼에러가 발생할 수 있다. 이를 방지하기 위해 어노테이터가 작업하는 단계를 1단계와 2단계로 나누어, 1단계에서 발생한 실수와 오류를 어노테이터에게 다시 교육하고 2단계 cross check 및 수정 단계를 거쳤다. 이를 통해 더 높은 퀄리티의 데이터셋을 구축할 수 있었다. 

Note that the construction process is divided into two phases because the raw text of \bookname is significantly long, where the average length of Korean text is 1,106 characters, and the history-specialized annotators are rare.
Before beginning the first phase, the annotators received instructions on the purpose of this study, the types of entities and relations, and how to operate the user interface (UI) for data tagging.
After instructions, annotators identified the named entities and the relations between them.
In the second phase, the annotators cross-checked the intermediate results and modified incorrect annotations.
During both phases, we provided the annotators with user guidance and maintained real-time communication.

\subsection{Corpus Collection}
As mentioned in \ref{sec:dataset_schema}, we selected 39 books from \bookname and divided them into 2,019 texts, each containing a single day's content.
We did not divide the text into shorter texts before providing it to the annotators
% When we give the text to the annotators, we do not split the text into the short texts.
because a relation may exist across multiple sentences or have its evidence sentence distant from where the relation appears. 
We provided the entire text to the annotators to reduce the possibility of losing relational data.
% To minimize the risk of loosing relational information, we deliver the whole text to the annotators.
Due to the highly variable length of the text, an additional process step was required to extract relational information in a manageable length.
To select the sentences containing all the information that can indicate the relational fact, we guided the annotators to detect the evidence sentence(s) when they annotated the relation types. 
% the context of near sentence

% 우리는 각 텍스트를 더 나누지 않았는데, 임의로 나눈 텍스트 사이에 관계가 존재할 수 있다고 판단하였기 때문이다.
% 하지만 각 텍스트의 길이가 매우 길고 길이가 매우 다양하기 때문에, 후처리로 관계를 훼손하지 않으면서 텍스트를 슬라이싱하기로 하였다. 따라서 어노테이터들에게 관계를 태깅할 때 그 근거문장 또한 태깅하도록 하였는데, 관계가 등장하는 문장과 근거문장을 하나의 긴 raw 텍스트에서 추출하기 위함이었다.

\subsection{Defining Schema}
\label{appendix-define-schema}
\subsubsection{Types of Named Entities}
As shown in Table~\ref{tab:entity_full}, we defined 10 entity types. 
Here, we added the date and time as entity type; thus, we can estimate the exact time because most of the corpus includes the time when the text was written.
For example, if a text contains tomorrow's plan by mentioning ``tomorrow'' and the written date is June 6, we can recognize the date of tomorrow as June 7.

In historical studies, it is essential to understand the lifestyle of ancient times.
Lifestyle includes clothing, food, and utilized products.
For instance, humans began consuming grains such as wheat and rice after the agricultural revolution.
Since lifestyle has changed according to time and location, detecting food, clothes, and products on our corpus becomes a non-trivial task.

We also excluded two text types in the preprocessing: poems and quotations.
When writing the \bookname, the writers commonly composed poems or quoted related or ancient books, including the Analects of Confucius and Mencius.
We decided to detect the books' name because it helps us imply the political status of the writer.
However, the poems usually describe the sentiments or thoughts of the writer, and the quotations are written in a more ancient time than Joseon. 
Since we concentrated on finding objective relational facts about the Joseon dynasty, we determined to exclude the poems and quotations.
A special ``exclude'' entity type was provided to the annotators, and the annotators tagged such subtexts if the text was a poem or a quotation.
% the part of the text if it was the poem or the quotation.

% \sy{table with 각 엔티티에 대한 설명}
% 기본이 되는 person, organization, location, 기타를 포함하는 others 을 포함하여, 총 9개의 객체를 지님. 이후 한문 관련 데이터셋에서 활용할 수 있을만한 객체를 선정함.
% - 날짜,시간: 책 제목이 날짜임. 그것을 기반으로 오늘, 내일이 며칠인지 유추할 수 있음.
% - 음식, 옷, 물건: 시대 혹은 장소에 따라 먹는 음식, 특산품, 사람/관직이 입는 옷, 사용하는 물건이 달라질 수 있음. 
% - 책 제목: 인용하는 책 제목의 변화를 통해 정치 경향의 변화를 알 수 있음.
% \sy{exclude class 설명. 제외할 부분}

\subsubsection{Types of Relations}
Since our corpus is a collection of travel reports, the authors wrote the people they had met and the places they had visited.
As shown in Table ~\ref{tab:relation_full}, we defined 20 relation classes, including 14 personal and 4 location relations.
% relations of person and four of location.
In the Joseon dynasty, it was a convention to refer to one another by their alternative name or title; % each other as alternate name or title the person hold. 
thus, identifying the alternative name of a specified person is essential for tracking the individual's life.
Also, since the name of a particular location can vary depending on time and place,
we added ``alternate name'' as a relation class to account for these instances.
Additionally, in \bookname, the number indicates the distance traveled from one location to another.
We hypothesized that the locations are close to each other if the text contains the distance between the locations where the author moved because there was no mechanical mobility and they usually walked the cities.
In addition, they described the characteristics of a location, such as its regional product or cuisine and its functional role.
% Also, they note the characteristics of a place, including the regional product or food and the functional role.
Therefore, ``loc:famous\_for'' and ``loc:function\_as'' were added to the set of relation types.

% \sy{table with description, percentage distribution}
% 사람과 관련된 관계는 N 개 이고, 장소와 관련된 관계는 K개이다. 
% 주로 사람, 지역, 기관, 물건 사이의 관계를 파악하고자 하였으며, others를 포함하여 총 20개의 관계 타입을 가짐.
% - 정보를 추출할 때 가장 어려운 점은 같은 대상을 가르치지만 다른 글자를 가지고 있는 '이칭'의 경우임. 특히 현대와 달리 고대 아시아에서는 사람을 사람 이름으로 부르지 않고 관직명이나 호를 불러 지식인이라면 사람 마다 호가 있었음. 따라서 한 사람에 대한 정보를 추출하는 데 있어 다른 이름을 태깅하는 것은 매우 중요함. 이에 'alternate\_name'이라는 관계를 만들었음. 이는 지명(location)에도 동일하게 적용됨. 실제로 'alternate\_name'의 경우 사람과 장소 사이에서 많이 태깅됨.
% - 한편 지명의 경우 현대 지명과 차이가 있어 과거의 지식과 현대의 지식을 잇기 어려운 점이 있음. 따라서 지명과 기관 사이에 nearby 라는 관계를 추가하여 지리적 지식을 추출하고자 함.
% - 사람이 어떤 관직을 수행하였는지, 가족 및 교우관계가 어떻게 되는지,
% - 정해진 루트를 다니는 연행록의 특성상 지명이 자주 등장함. 이때 해당 장소의 특산품이나 역할을 소개하곤 하는데, 이를 태깅하기 위해 loc:famous\_for, loc:function\_as를 추가함.

\subsection{Entity Detection}
\label{sec:entity_detection}
% \sy{UI 넣기. parallel relationship vs. relation type.}

The annotators annotated entities using a predefined set of entity types. 
We provided the original Hanja and the translated Korean texts, as shown in Fig.~\ref{fig:UI}.
As most annotators' native language is Korean, we recommended detecting the entities in the Korean text first and the parallel entities in the Hanja text after.
After detecting entities in both texts, the annotators drew a line connecting the same entity between the two languages (as in \textit{apple} and \textit{pomme} in English and French texts).
The annotators also drew a line connecting entities that express a certain relation.
To avoid confusion, the two lines are colored in blue and orange, respectively, as shown in Figure~\ref{fig:UI}.
% Note that these parallel relations exist between an entity written in the source language, i.e., Hanja, and the same entity written in the target language, i.e., Korean 
% To avoid confusion between the parallel relation and relation triplets between different named entities, we colored the parallel line blue and the relation line orange, as shown in Fig.~\ref{fig:UI}.
% After detecting the named entity and parallel relationship, the annotators check all entities are detected.

% \sy{entity, parallel, co-reference type}
% 어노테이터들은 사전에 정의된 객체에 대응되는 객체가 등장할 경우 이를 태깅하였다. 다만 텍스트가 한문과 이를 번역한 한국어 텍스트가 주어지는데, 어노테이터 대부분이 한국인이었으므로 한국어 텍스트에 대하여 먼저 태깅을 진행한 후 이에 대응되는 한문 객체를 태깅하고 동일한 대상을 가르키는 한국어 객체와 한문 객체 사이에 parallel 관계를 태깅하도록 하였다. 
% parallel 과 일반 관계의 구분을 편하게 하기 위해 UI 상에서 두 관계의 태깅 결과를 구분하여 시각화하였다. 
% 또한 병렬 관계가 알맞게 태깅되었는지를 보기 편하게 파란 선으로 그 관계를 표기하였으며, 일반 관계의 경우 주황색으로 표기하여 구분이 쉽게 하였다.
% 하나의 텍스트에 대하여 엔티티를 태깅한 뒤 텍스트에서 등장한 객체가 모두 태깅 되었는지를 확인한 뒤 관계를 태깅하였다.

\subsection{Relation Annotation}
\label{sec:relation_annotation}
% Since the relation was annotated between the named entities in the previous step and the parallel relation was added in the Korean and Hanja text, 
After identifying the relations in the previous step, the annotators added relations by using the ``add relation'' button and selected a relation class for the relation triplet.
They also tagged the indices of evidence sentences on the Korean and Hanja texts.
% , where cannot obtain the parallel relationship at the sentence level.
% Thanks to the annotators' elaborate result, we can ex

% \sy{relation, evidence sentence on Korean and Hanja}
% 한국어와 한문의 모든 객체에 대하여 parallel 관계가 태깅되어있으므로, 관계의 경우 한국어 텍스트에서만 진행하였고 한문의 관계는 parallel 관계를 사용하여 후처리 단계에서 추가하였다. 예를 들어 A\_kor, B\_kor 사이에 관계가 존재한다면 parallel 로 대응되는 한문 객체인 A\_han 과 B\_han 사이에 동일한 관계를 태깅하도록 하였다. 
% 관계는 UI에서 객체를 태깅한 뒤 +add relation 버튼을 클릭한 뒤 tail에 대응하는 객체를 클릭하면 관계를 생성할 수 있었고, 사전에 정의된 관계 클래스 중 적합한 것을 선택하도록 하였다.
% 또한 evidence sentence를 텍스트에 보이는 문장의 인덱스로 추가하도록 하였다. 이 때 근거 문장은 한국어와 한문 각각에 대하여 태깅하였는데, 관계 정보는 parallel 정보에서 추출할 수 있지만 문장간의 parallel 정보는 따로 얻을 수 없기 때문이다. 

\subsection{Cross-Checking and Modification}
After the first phase, we analyzed the intermediate result and updated the user manual, focusing on instructions for editing initial annotations.
Before the cross-checking stage, we conducted a second tutorial for the annotators using the updated manual. % the second education.
We assigned annotators to texts such that they had not seen them during the first phase.
% the text to annotators, different from the text dealt with in the first phase.
If they found an error(s) during cross-checking, they revised the annotations by adding or removing the entity(s) or relation(s).
% 앞서 작업한 결과물을 작업 관리자가 검토한 뒤 어노테이터들에게 2차 교육을 진행하여 오류 사항을 일괄적으로 수정할 수 있도록 하였다. 어노테이터는 1차에서 작업한 텍스트가 아닌, 다른 어노테이터가 작업한 작업물을 태스크로 받고 검수 및 수정 작업을 진행하였다. 

\section{Experiments}
% \label{sec:appendix_exp}
% In this section, we will add computational details following the Responsible NLP checklist.

\subsection{Computational Details}
\label{sec:appendix_exp_compu}

% \sy{number of parameters, total computational budget (hours)}
Our experiments include monolingual and bilingual settings. For each model, we describe the number of total parameters and computational budget (hours) for training on 200 epochs on our dataset when \SL is 0.
For the Korean model, mBERT consists of 178M parameters and consumes about 4.2 hours, KoBERT is 93M and 3.3 hours, and KLUE is 111M and 4.0 hours, respectively. 
For the Hanja model, mBERT consists of 178M parameters and requires 4.6 hours, and AnchiBERT is 95M and 3.3 hours.
Our joint model consists of 206M parameters and consumes 6.6 hours because our model adopts two separate PLMs.

\begin{table*}[t!]
\centering
\resizebox{\textwidth}{!}{%
\begin{tabular}{ll|ccc|ccc|ccc}
    \toprule
    \multicolumn{2}{l|}{} & \multicolumn{3}{c|}{\textit{SL} = 2} & \multicolumn{3}{c|}{\textit{SL} = 4} & \multicolumn{3}{c}{\textit{SL} = 8} \\
    Language & Model & P & R & F1 & P & R & F1 & P & R & F1 \\  \hhline{==|======|===} 
    \multirow{3}{*}{Korean} 
    & mBERT & 57.43 & 42.69 & 48.97 & 37.15 & 38.80 & {\ul 37.96} & 18.16 & 20.86 & 19.41 \\
    & KoBERT &  47.01 & 31.43 & 37.67 & 14.54 & 14.32 & 14.43 & 7.35 & 5.46 &  6.27 \\
    & KLUE & 54.93 & 45.47 & {\ul 49.75} & 36.36 & 38.21 & 37.27 & 16.76 & 25.54 & {\ul 20.24} \\
     \hline
    \multirow{2}{*}{Hanja}
    & mBERT &  26.81 & 26.24 & 26.52 & 17.58 & 18.73 & 18.14 & 9.58 & 13.69  &  11.27 \\
    & AnchiBERT &  32.27 & 32.12 & {\ul 32.24} & 22.11 & 22.87 & {\ul 22.48 } & 15.16 & 18.71 & {\ul 16.75 } \\
     \hline
    % \multirow{2}{*}{Korean+Hanja}
     Korean+Hanja & Ours &  66.73 & 41.24 & \textbf{50.98} & 48.27 & 36.21 & \textbf{41.38} & 25.30 & 21.97 & \textbf{23.52}\\
     \bottomrule
\end{tabular}%
}
\caption{
Performance comparison when \SL is 2, 4, and 8.
P, R, and F1 are precision, recall, and F1 score respectively.
All scores are described on the percentage (\%) and rounded off the third decimal point.
The \textbf{best F1 score} is in bold at each \textit{SL}, and the \underline{second} score for each language is underlined.
}
\label{tab:performance-sl-48}
\end{table*}
\label{sec:appendix_exp_add_exp}

\subsection{Performance Comparison on Large \SL}
As shown in Table~\ref{tab:performance-sl-48}, our joint model outperforms other baseline models when \SL is 2, 4, and 8, where the average length of documents is 153, 250, and 427 tokens on the Korean text.
Our model scores better when $\alpha$ is 0.6 rather than 0.5 when \SL is 2, 4, and 8. 
This can be explained by the fact that ours is affected by the low performance of the Hanja encoder, i.e., AnchiBERT. The Hanja encoder significantly drops its scores as \SL increases.
% Therefore, our model shows the best score when the model gives additional weight to the predictions of the Korean encoder. 

\section{Dataset Examples}

We include additional full data samples: Table~\ref{tab:appendix_exp3}, Table~\ref{tab:appendix_exp1}, and Table~\ref{tab:appendix_exp2}.

% as well as the data samples of Fig.~\ref{fig:case_study} in Table~\ref{tab:appendix_exp1} and Table~\ref{tab:appendix_exp2}. Since the model analysis is calculated when SL is 4, the document becomes extremely long. Therefore, we edit several sentences in each table.  
% \sy{dataset example with Eng. translated text}

% \usepackage{graphicx}
\begin{table*}[ht]
\centering
\resizebox{\textwidth}{!}{%
    \begin{tabular}{lrrl}
    \toprule
    \textbf{Entity type} & \multicolumn{1}{c}{\textbf{Frequency}} & \multicolumn{1}{c}{\textbf{Ratio (\%)}} & \multicolumn{1}{c}{\textbf{Description}}  \\ \midrule
    Person & 22,998 & 34.55 & People, the alternate name of a specific person, title \\ 
    Location & 23,900 & 35.91 & \begin{tabular}[c]{@{}l@{}}Geogprahically defined locations, including mountains and waters, etc. \\ Politically defined locations, including countries, cities, states, etc. \\ Facilities, including building, etc. \end{tabular}  \\ 
    Organization & 1,806 & 2.71 & Institutions, political or religious groups, etc. \\  
    Number & 9,057 & 13.61 & Money and quantities, including distance between locations, etc.  \\  
    Datetime & 3,210 & 4.82 & Absolute or relative dates, times, or periods. \\ 
    Product & 2,927 & 4.40 & Gifts, regional specialties, tributes, and animal, etc. \\ 
    Food & 550 & 0.83 & Meal, snack, fruits, and drinks, etc.  \\
    Clothes & 753 & 1.13 &  Garment or dress. \\
    Book & 287 & 0.43 & Antique or referred name of books \\
    Other & 1,068 & 1.60 & Relevant entity type which are not included in the predefined types. \\ \hline
    \textbf{Total} & 66,556 & 100.00 & \\
    \bottomrule
    \end{tabular}%
}
\caption{List of entity types.}
\label{tab:entity_full}
\end{table*}

% \usepackage{graphicx}

%  \begin{tabular}[c]{@{}l@{}}Geogprahically defined locations, including mountains and waters, etc. \\ Politically defined locations, including countries, cities, states, etc. \\ Facilities, including building, etc. \end{tabular}

\begin{table*}[ht]
\centering
\resizebox{\textwidth}{!}{%
\begin{tabular}{lrrl}
\toprule
\textbf{Relation type} & \multicolumn{1}{c}{\textbf{Frequency}} & \multicolumn{1}{c}{\textbf{Ratio (\%)}} & \multicolumn{1}{c}{\textbf{Description}}  \\ \midrule
nearby & 2,718 & 27.28 & \begin{tabular}[c]{@{}l@{}}  The location or organization are geographically close to the specified \\ location or organization.\end{tabular} \\
alternate\_name & 756 & 7.59 & \begin{tabular}[c]{@{}l@{}} Alternative names called instead of the official name to refer the \\ specified person, organization, location, etc.\end{tabular} \\ 
per:position\_held  & 3,194 & 32.05 & Title that represent the position of the specified person. \\
per:worn\_by  & 353 & 3.54 & Garment or dress that the specified person wears. \\
per:friend  & 143 & 1.44 & The friend of the specified person \\
per:enemy  & 49  & 0.49 & The person or organization that the specified person is hostile to.\\
per:child  & 113 & 1.13 & The children of the specified person. \\
per:sibling  & 75 & 0.75 &  The brothers or sisters of the specified person.\\
per:other\_family  &  168  & 1.69  & \begin{tabular}[c]{@{}l@{}} Family members of the specified person other than parents, children, \\ siblings. \end{tabular} \\
per:country\_of\_citizenship  & 533  & 5.35  & The nationality of the specified person. \\
per:place\_of\_residence  & 364  & 3.65 & The place where the specified person lives. \\
per:place\_of\_birth  & 58 & 0.58 & The place where the specified person was born. \\
per:place\_of\_death & 26 & 0.26 &  The place where the specified person died.\\
per:date\_of\_birth & 10 & 0.10 &  The date when the specified person was born.\\
per:date\_of\_death & 8 & 0.08 &  The date when the specified person was died. \\
loc:functions\_as & 319  & 3.20 & The political or functional role of the specified location.  \\
loc:famous\_for & 64 & 0.64 & The regional product or food that is famous at the specified location.  \\
product:provided\_by & 381 & 3.82 & The organization or person that gives the specified product.  \\
org:member\_of & 369 & 3.70 & The specified person who belongs to the specified organization. \\
others & 264 & 2.65 & Relevant relation class which are not included in the predefined classes. \\ \hline
\textbf{Total} & 9,965 & 100.00 & \\
\bottomrule
\end{tabular}%
}
\caption{List of relation types.}
\label{tab:relation_full}
\end{table*}


\begin{table*}[ht]
\centering
\resizebox{\textwidth}{!}{%
\begin{tabular}{l|l}
\toprule
Text\_Kor & \begin{tabular}[c]{@{}l@{}} \Location{성안} 좌우에 벌여 있는 \Location{전사}는 모양이 \Location{우리나라}와 같고 큰길도 \Location{우리나라} 길보다 넓지 않았으나 길가에 원래 가가짓는 규례가 없다.\\ 일찍이 들으니 입성하는 날은 거마 때문에 길이 막혀서 전진하기가 어렵다 하더니,이번은 일행이 쌍쌍으로 어깨를 나란히 하고 임의대로 갔으며 \\ 좌우로 눈에 보이는 것도 \Location{통주}보다 나을 것이 없다. 길에서 \Clothes{누런 비단 모자}에 \Clothes{누런 비단 옷}을 입은 자를 만났다. 괴이쩍어서 물었더니, \\ \Location{황제의 원찰}에 있는 \Person{몽고 승려}라 답하였다. 입성한 후에 왕래하는 여인은 모두 호녀였으며 저자에 출입하는 계집은 없었다. \end{tabular}  \\ \hline
Text\_Han &  \begin{tabular}[c]{@{}l@{}} 第\Location{城中}左右\Location{廛舍}. 狀如\Location{我東}. 而大路亦不廣於\Location{我國}. 而第路邊元無結假家之規. 曾聞入城之日. \\ 礙於車馬. 實難前進矣. 今則一行雙雙比肩. 任意作行. 而左右耳目之所睹. 決不過於\Location{通州}. \\ 路逢着\Clothes{黃錦帽}\Clothes{黃錦衣}者. 怪而問之. 則答云\Location{皇帝願堂寺}\Person{蒙古僧}也. \end{tabular}  \\ \hline
Text\_Eng* & \begin{tabular}[c]{@{}l@{}} 
    \Location{The temple} on the left and right sides of \Location{the fortress} has the same shape as \Location{Korea}, and the main road was not wider than that of \Location{Korea}, \\ but there is no original rule on the side of the road. I heard earlier that it was difficult to move forward on the day of entering the country \\ because the road was blocked due to the kiln, but this time, the party went arbitrarily, shoulder to shoulder in pairs, \\ and what is visible to the left and right is no better than \Location{Tongju}. I met a man in \Clothes{a yellow silk hat} and \Clothes{a yellow silk dress} on the street. \\ When I asked him in a strange way, he replied that he was \Person{a Mongolian monk} in \Location{the emperor's original temple}. \\ All the women who came and went after entering the country were women, and there were no women who entered the author.
\end{tabular}  \\ \midrule
Entity & \Location{Location}, \Person{Person}, \Clothes{Clothes}  \\ \hline
Relation & \begin{tabular}[c]{@{}l@{}}
    (`sbj\_kor': 몽고 승려, `sbj\_han': 蒙古僧, `obj\_kor': 누런 비단 옷, `obj\_han': 黃錦衣, `relation': per:worn\_by), \\
    (`sbj\_kor': 몽고 승려, `sbj\_han': 蒙古僧, `obj\_kor': 누런 비단 모자, `obj\_han': 黃錦帽, `relation': per:worn\_by)\end{tabular} \\ \hline
Meta data & \begin{tabular}[c]{@{}l@{}}
    `book\_title': 연행록, `text\_chapter': 임진년(1712, 숙종 38) 12월, `title': 27일 (3), `writer': 최덕중, `year': 1712, \\
    `book\_volume': 일기(日記), `copyright': ⓒ 한국고전번역원 | 이익성 (역) | 1976
\end{tabular} \\
\bottomrule
\end{tabular}%
}
\caption{\datasetname example when \textit{SL}=2.}
\label{tab:appendix_exp3}
\end{table*}

\begin{table*}[ht]
\centering
\resizebox{\textwidth}{!}{%
\begin{tabular}{l|l}
\toprule
Text\_Kor & \begin{tabular}[c]{@{}l@{}}마을 집이 물 양쪽 언덕에 갈라 있어서 지형과 마을 제도가 \Location{십리보} 마을과 같았다. \Location{사하보}에서 \Number{5리쯤} 거리에 \Location{포교와촌}이 있고 \\ \Location{포교와촌}에서 \Number{8리}쯤 거리에 \Location{화소교}ㆍ\Location{전장포} 등 마을이 있었다. \Location{백탑보}에서 \Number{10여 리}를 가니 \Location{혼하}가 있는데, 일명 \Location{아리강}이다. \\ \Location{아리강} 남쪽 언덕에 \Person{관장} 3형제의 기마상이 있었다. 강변에 나룻배와 마상선이 있었다.\end{tabular}  \\ \hline
Text\_Han &  \begin{tabular}[c]{@{}l@{}}如\Location{十里堡}之村居. \Location{堡}去\Number{五里}許. 有\Location{暴交哇村}. \Location{村}去\Number{八里}許. 有\Location{火燒橋},\Location{氊匠鋪}等村矣. \\ 自\Location{白塔堡}行\Number{十餘里}. 有\Location{混河}. 而一名\Location{阿利江}. \Location{江}之南岸. 有\Person{關將}三昆季騎馬之像. 江邊有津船及馬上船.\end{tabular}  \\ \hline
Text\_Eng* & \begin{tabular}[c]{@{}l@{}}The village house was divided on both sides of the water, so the topography and village system were the same as \Location{Sipribo} Village. \\ Pogyo Village was located about \Number{5 ri} away from \Location{Sahabo}, and there were villages such as \Location{Hwasogyo Bridge} and \Location{Jeonjangpo} \Number{8 ri} away \\ from \Location{Pogyo Village}. After going about \Number{10 ri} from \Location{Baektapbo}, there is \Location{Honha}, also known as \Location{Arigang}.  On the southern hill of the \Location{Ari River}, \\  there was a mounted statue of the three \Person{officers}. There were ferry boats and horseboats along the river.\end{tabular}  \\ \midrule
Entity & \Location{Location}, \Person{Person}, \Number{Number}  \\ \hline
Relation & \begin{tabular}[c]{@{}l@{}}(`sbj\_kor':혼하 , `sbj\_han': 混河, `obj\_kor': 아리강, `obj\_han': 阿利江, `relation': alternate\_name), \\ (`sbj\_kor': 백탑보, `sbj\_han': 白塔堡, `obj\_kor': 혼하, `obj\_han': 混河, `relation': nearby )\end{tabular} \\ \hline
Meta data & \begin{tabular}[c]{@{}l@{}}`book\_title': 연행록, `text\_chapter': 임진년(1712, 숙종 38) 12월, `title': 6일 (3), `writer': 최덕중, `year': 1712,\\ `book\_volume': 일기(日記), `copyright': ⓒ 한국고전번역원 | 이익성 (역) | 1976
\end{tabular} \\
\bottomrule
\end{tabular}%
}
\caption{\datasetname example when \textit{SL}=2.}
\label{tab:appendix_exp1}
\end{table*}

\begin{table*}[ht]
\centering
\resizebox{\textwidth}{!}{%
\begin{tabular}{l|l}
\toprule
Text\_Kor & \begin{tabular}[c]{@{}l@{}}이는 만일 \Location{우리나라}의 \Person{별사}가 동시에 입성하게 되면, 또한 \Location{관}을 \Location{북문} 안에 설치하는 까닭에 \Location{남관}ㆍ\Location{북관}으로 구별하게 된 것이다. \\ 관은 대개 \Number{100여 칸}인데 가로 세로가 모두 일자 모양으로 되었으며, 관문 안에 \Location{중문}이 있고 \Location{중문} 안에 동서로 \Location{낭옥}이 있는데, \\ 이것은 \Location{원역의 무리들이 거처하는 곳}이다. 또 \Location{소문} 안에 \Location{정당}이 있는데 \Location{정사가 거처하는 곳}이며 그 좌우 \Location{월랑의 상방}은 \Location{편막들이 거처하는 곳}이었다. \\ 또 북쪽으로 제2, 제3의 집에는 \Person{부사}와 \Person{서장관}이 나누어 거처하고, \Location{편막}들 역시 본 방의 곁채에 나누어 들었다. 뒤쪽에 온돌 \Number{십수 칸}이 있어, \\ \Person{원역}ㆍ하인ㆍ\Product{말}들이 그 속에 함께 들었는데, 수숫대로 엮고 연지로 발라 각각 칸막이를 하였다. \end{tabular}  \\ \hline
Text\_Han &  \begin{tabular}[c]{@{}l@{}} 若\Location{我國}\Person{別使}同時入城. 則又設一\Location{館}於\Location{北門}內. 故有南北\Location{館}之別也. 館凡\Number{百餘間}. 皆縱橫爲一字制. \Location{館門}內有\Location{中門}. \\ \Location{中門}內有東西\Location{廊屋}. 此員譯輩所處也. 又於\Location{小門}內有\Location{正堂}. \Location{正使處}焉. 左右\Location{月廊上房}. \Location{褊幕所處}也. \\ 又北而第二第三行則 \Person{副使}, \Person{書狀}分處焉. \Location{褊幕}則亦分入本房夾廊. 後邊有北炕\Number{十數間}. \Person{員譯}及下輩人\Product{馬}. \end{tabular}  \\ \hline
Text\_Eng* & \begin{tabular}[c]{@{}l@{}}This is because if a \Location{Korean} \Person{monk} enters at the same time, \Location{the coffin} was also installed inside \Location{the north gate} and it was distinguished as \Location{Namgwan} \\ and \Location{Bukgwan}. The coffin is usually about \Number{100 compartments}, all of which are straight in width and length, and there is \Location{a middle gate} \\ inside the gate and a \Location{Nangok} from east to west inside \Location{the middle gate}, which is a place \Location{where groups of original stations live}. \\ Also, there is a \Location{Jeongdang}, \Location{where Jeongsa lives}, and the left and right \Location{Wollang} was \Location{where the Pyeonak lived}. \\ In addition, in the second and third houses to the north,  \Person{the deputy} and  \Person{the minister Seo} lived separately, and the  \Location{Pyeonmak} were also divided into \\ the side quarters of the main room. There was an ondol \Number{ten-square compartment} in the back, and \Person{the original station}, servants, \\ and \Product{horses} were included in it, and they were woven with a sorghum stick and applied with rouge to separate them.\end{tabular}  \\ \midrule
Entity & \Location{Location}, \Person{Person}, \Product{Product}  \\ \hline
Relation & \begin{tabular}[c]{@{}l@{}}
    (`sbj\_kor':소문 , `sbj\_han': 小門, `obj\_kor': 정당, `obj\_han': 正堂, `relation': nearby), \\
    (`sbj\_kor':정당 , `sbj\_han': 正堂, `obj\_kor': 정사가 거처하는 곳, `obj\_han': 正使處, `relation': loc:functions\_as), \\
    (`sbj\_kor': 월랑의 상방, `sbj\_han': 月廊上房, `obj\_kor': 편막들이 거처하는 곳, `obj\_han': 褊幕所處, `relation': loc:functions\_as )\end{tabular} \\ \hline
Meta data & \begin{tabular}[c]{@{}l@{}}
    `book\_title': 계산기정, `text\_chapter': 도만(渡灣) ○ 계해년(1803, 순조 3) 12월[4일-24일], `title': 24일(을유) (2), `writer': '미정', `year': 1803 \\  `book\_volume': 계산기정 제2권, `copyright': ⓒ 한국고전번역원 | 차주환 (역) | 1976 
\end{tabular} \\
\bottomrule
\end{tabular}%
}
\caption{\datasetname example when \textit{SL}=2.}
\label{tab:appendix_exp2}
\end{table*}

% \begin{table*}[]
% \centering
% \resizebox{\textwidth}{!}{%
% \begin{tabular}{l|l}
% \toprule
% Text\_Kor & \begin{tabular}[c]{@{}l@{}} \end{tabular}  \\ \hline
% Text\_Han &  \begin{tabular}[c]{@{}l@{}} \end{tabular}  \\ \hline
% Text\_Eng* & \begin{tabular}[c]{@{}l@{}} \end{tabular}  \\ \midrule
% Entity & \Location{Location}, \Person{Person}, \Product{Product}  \\ \hline
% Relation & \begin{tabular}[c]{@{}l@{}}
%     ('sbj\_kor':소문 , 'sbj\_han': 小門, 'obj\_kor': 정당, 'obj\_han': 正堂, 'relation': nearby), \\
%     ('sbj\_kor': 월랑의 상방, 'sbj\_han': 月廊上房, 'obj\_kor': 편막들이 거처하는 곳, 'obj\_han': 褊幕所處, 'relation': loc:functions\_as )\end{tabular} \\ \hline
% Meta data & \begin{tabular}[c]{@{}l@{}}
%     'book\_title': '계산기정', 'text\_chapter': '도만(渡灣) ○ 계해년(1803, 순조 3) 12월[4일-24일]', 'title': '24일(을유) (2)', 'writer': '미정', 'year': 1803 \\  'book\_volume': '계산기정 제2권', 'copyright': 'ⓒ 한국고전번역원 | 차주환 (역) | 1976', 
%     % 'book\_title': '연행록', 'text\_chapter': '임진년(1712, 숙종 38) 12월', 'title': '6일 (3)', 'writer': '최덕중', 'year': 1712,\\ 'book\_volume': '일기(日記)', 'copyright': 'ⓒ 한국고전번역원 | 이익성 (역) | 1976'
% \end{tabular} \\
% \bottomrule
% \end{tabular}%
% }
% \caption{Example of \datasetname when \textit{SL} is .}
% \label{tab:appendix_exp2}
% \end{table*}

% \end{document}

\end{document}